\documentclass[journal]{IEEEtran}
\usepackage{comment}
\usepackage{graphicx}
\usepackage{amsmath}
\usepackage{amsfonts}
\usepackage{algorithm}
\usepackage{algorithmic}
\usepackage{makecell}
\usepackage{booktabs}
\setcellgapes{3pt}
\usepackage{multirow}
\usepackage{color}

\setcellgapes{3pt}

\usepackage[justification=centering]{caption}
\usepackage{color}

\usepackage{cite}

\ifCLASSINFOpdf
\else
\fi
\begin{document}
%
\title{Stacked Auto Encoder Based Deep Reinforcement Learning for Online Resource Scheduling in Large-Scale MEC Networks }

\author{Feibo Jiang, Kezhi Wang, Li Dong, Cunhua Pan and Kun Yang
	\thanks{

This work was supported in part by the National Natural Science Foundation
of China under Grant no. 41604117, 41904127, 41874148, 61701179, 61620106011 and 61572389. This work was also supported in part by Scientific Research Fund of Hunan Provincial Education Department in China under Grant no. 18A031, and supported in part by the Hunan Provincial Science Technology Project Foundation under Grant no. 2018TP1018 and 2018RS3065, and supported in part by UK EPSRC Project NIRVANA (EP/L026031/1)

Feibo Jiang (jiangfb@hunnu.edu.cn) is with Hunan Provincial Key Laboratory of Intelligent Computing and Language Information Processing, Hunan Normal University, Changsha, China, Kezhi Wang (kezhi.wang@northumbria.ac.uk) is with the department of Computer and Information Sciences, Northumbria University, UK, Li Dong (Dlj2017@hunnu.edu.cn) is with Key Laboratory of Hunan Province for New Retail Virtual Reality Technology, Hunan University of Technology and Business, Changsha, China, Cunhua Pan (Email: c.pan@qmul.ac.uk) is with School of Electronic Engineering and Computer Science, Queen Mary University of London, London, E1 4NS, UK, Kun Yang (kunyang@essex.ac.uk) is with the School of Computer Technology and Engineering, Changchun Institute of Technology, Changchun, China and also with the School of Computer Sciences and Electrical Engineering, University of Essex, CO4 3SQ, Colchester, UK.
	
Corresponding authors: Kezhi Wang; Li Dong.}
}

\markboth{Submitted for Review}%
{Shell \MakeLowercase{\textit{et al.}}: Bare Demo of IEEEtran.cls for IEEE Journals}
%



\maketitle

\begin{abstract}

An online resource scheduling framework is proposed for minimizing the sum of weighted task latency for all the Internet of things (IoT) users, by optimizing offloading decision, transmission power and resource allocation in the large-scale mobile edge computing (MEC) system. Towards this end, a deep reinforcement learning (DRL) based solution is proposed, which includes the following components. Firstly, a related and regularized stacked auto encoder (2r-SAE) with unsupervised learning is applied to perform data compression and representation for high dimensional channel quality information (CQI) data, which can reduce the state space for DRL. Secondly, we present an adaptive simulated annealing based approach (ASA) as the action search method of DRL, in which an adaptive h-mutation is used to guide the search direction and an adaptive iteration is proposed to enhance the search efficiency during the DRL process. Thirdly, a preserved and prioritized experience replay (2p-ER) is introduced to assist the DRL to train the policy network and find the optimal offloading policy. Numerical results are provided to demonstrate that the proposed algorithm can achieve near-optimal performance while significantly decreasing the computational time compared with existing benchmarks.

\end{abstract}

\begin{IEEEkeywords}
Stacked auto encoder, deep reinforcement learning, adaptive simulated annealing, large-scale mobile edge computing.
\end{IEEEkeywords}

%
\IEEEpeerreviewmaketitle

\section{Introduction}

%
%
%
%
In recent years, the number of user equipments (UEs), e.g., mobile phones and Internet of Things (IoT) devices are growing rapidly. Meanwhile, many new resource-intensive applications, e.g., augmented reality (AR), virtual reality (VR), real-time gaming, face recognition and natural language processing are constantly emerging.  
However, the above attractive applications normally require large amount of computing resource and are latency-sensitive. The UEs, due to its limited size and resource, may not be able to complete the above tasks in required time or meeting the Quality of Service (QoS) requirement.

Mobile edge computing (MEC) is proposed to enable UEs to offload the above-mentioned tasks to edge servers and has attracted much attention from both academia and industry \cite{mao2017survey}. There are two main advantages of applying MEC to assist the UEs. Firstly, the local energy consumption of UE may be reduced as the UE can offload the computation-intensive tasks to the MEC. Secondly, the response time could be decreased, as the MEC normally has much more computing resource than the local device and therefore could complete tasks much faster than the local device and thereby increasing the user experience significantly. However, when we have a large scale of users, one MEC may not be powerful enough and thus multiple MECs could be deployed. Then, the key question here is that how we can decide the user association and resource allocation, especially in large-scale environment \cite{mao2017survey, 8754787,tang2019waiting}. 

 In addition, several works have been proposed to optimize the 	latency-sensitive services, i.e., virtual reality applications, over mobile edge computing \cite{8932591, 8491367, 8387798, 8789664}. Reference \cite{8932591} proposed the secured offloading optimization framework for low-latency MEC systems. The MEC with caching assisted low-latency system was studied in \cite{8491367}. Also, latency optimization for resource allocation has been studied in \cite{8387798}. Moreover, reference \cite{8789664} proposed the latency optimal task assignment and resource allocation for heterogeneous multi-layer MEC systems.

The above problem is generally considered to be a mixed-integer non-linear programming (MINLP), as the offloading decision is always the integer variables whereas the resource allocation are the continuous variables. Some traditional methods were proposed to solve the above MINLP problem, such as dynamic programming\cite{bertsekas1995dynamic}, branch-and-bound method\cite{narendra1977branch} and game theory\cite{liu2017decentralized}.
However, these methods normally have high computational complexity, especially in large-scale scenarios. Also, some heuristic search\cite{bi2018computation} and convex based relaxation\cite{dinh2017offloading} were proposed, but these algorithms normally need several iterations to converge and therefore may not be suitable for fast decision making process. In the multi-MEC system with multi-user scenarios, the time-varying wireless channel largely impacts the optimal decision making process, which is very challenging for the above-mentioned traditional algorithms to deal with, as those traditional solutions normally requires to re-run the algorithms, once the environment changes. 

Fortunately, machine learning (ML) based solutions show great potential in addressing the above-mentioned issues by applying adaptive modelling and intelligent learning. Once the training is completed, normally the solutions can be obtained quite fast as only a few number of algebra calculations are needed. 
Recently, some ML or deep learning (DL) based algorithms have been proposed and applied to MEC systems, such as DNN\cite{huang2018distributed}, LSTM\cite{jiang2018predicted}, CNN\cite{liu2019energy}, Q-learning\cite{xiao2016mobile}, DQN\cite{he2017integrated} and DDPG\cite{liu2018energy}. However, on one hand, the DL-based models (e.g. DNN, LSTM and CNN) have outstanding prediction and reasoning capabilities, but they require considerable amount of labelled training data\cite{jiang2018electrical,jiang2018using,li2018learning}. On the other hand, when the scale of the MEC system grows, reinforcement learning (RL)-based models (e.g. Q-learning, DQN and DDPG) are not able to converge and the final results are unstable\cite{huang2019deep,henderson2018deep,10.1145/3234463} . 

 Against the above background, in this paper, we propose a comprehensive framework to jointly optimize computation offloading and resource allocation in the large-scale MEC system with multiple UEs deployed. We aim to obtain an online scheduling algorithm to minimize the sum of weighted task latency for all the UEs. Towards this end, we propose a DRL based framework with the following three components, i.e., related and regularized stacked auto encoder (2r-SAE), adaptive simulated annealing approach (ASA), and preserved and prioritized experience replay (2p-ER). Compared with the existing works, we have the following contributions:

Firstly, we propose a 2r-SAE with unsupervised learning to carry out data compression and representation for high-dimensional channel quality data. 2r-SAE can provide a compact data representation to the DRL model, which can reduce the state space and enhance the learning efficiency of the DRL. In addition, we add the relative error term of each UE to the error term of the loss function, which can consider the relative and absolute error simultaneously and reduce the information loss of each UE in the feature extraction process. We also add a regularization term to the loss function to improve the generalization of SAE. Furthermore, the incremental learning is used to update the SAE for tracking the variations of the real scenarios.

Then, we present an ASA approach as the heuristic search method to find the optimal action for the DRL model to generate offloading decision
with the corresponding state. In the ASA, we introduce two adaptive mechanisms: On one hand, the subsequent solution is mutated adaptively according to the channel quality information. On the other hand, the iteration number is adjusted adaptively according to the loss decrease of DRL. These two mechanisms can enhance the efficiency of SA and reduce the time of solving the original optimization without compromising the system performance.

Finally, a 2p-ER method is proposed to train the deep neural network (DNN) in DRL framework. In particular, we use a preserve strategy to protect the transitions which are close to the current offloading policy. We also adopt a priority strategy to select the transitions which can make more contributions to the decrease of loss function. These two strategies can accelerate the convergence of the DRL, which are important for large-scale networks.

The rest of this paper is organized as follows. In Section II, a review of related works is presented. Then, we describe the system model and problem formulation in Section III. We introduce the detailed designs of the DRL framework in Section IV. Section V provides the numerical results, followed by the conclusions in Section VI.


 

\section{Related works}
There were some contributions in the MEC systems applying artificial intelligence (AI)-based solutions. In the following, we review the related works from three aspects:  DL-based methods, RL-based methods and other AI-based methods.

DL-based methods: In \cite{huang2018distributed}, a distributed DL algorithm was proposed to make offloading decision for MEC systems, where several DNNs were trained parallelly and the offloading decisions were made cooperatively. In \cite{jiang2018predicted}, a long and short-term memory (LSTM) network was proposed to predict the traffic of small base stations (SBSs), and the cross-entropy loss function was applied to evaluate the LSTM and obtain the offloading strategy. In \cite{li2018learning}, a distributed deployment strategy for the multi-layer convolutional neural network was presented, which included two parts: the preprocessing part and the classification part. The preprocessing part was deployed on the edge server for feature extraction and data compression so as to reduce the data transmission between the edge and cloud system.

RL-based methods: In \cite{xiao2016mobile}, Q-learning-based mobile offloading strategy was proposed in the mobile offloading game. In \cite{he2017integrated}, a DQN based approach was applied to jointly optimize the networking, caching, and computing resources in the vehicular networks. In \cite{liu2018energy}, a DRL-based energy-efficient UAV control method was proposed to design the trajectory of UAV by jointly considering the communications coverage, fairness, energy consumption and connectivity.

Other AI-based methods: In \cite{guo2018efficient}, the energy-efficient computation offloading management scheme in the MEC system with small cell networks (SCNs) was proposed, and a hierarchical genetic algorithm (GA) and particle swarm optimization (PSO)-based heuristic algorithm were designed to solve this problem. In \cite{chen2017caching}, a conceptor-based echo state network was proposed to predict content request distribution of users and its mobility pattern when the network is available. Based on the prediction results, the optimal positions of UAVs and the content to cache at UAVs can be obtained.

However, none of above works consider the online decision making and resource allocation for large-scale MEC systems in dynamic environment. Firstly, DL-based methods need prior knowledge and labelled samples, which may be hard to obtain for the dynamic environment. Secondly, RL-based methods may be unstable and hard to converge for large search space with large-scale users. Thus, more flexible and efficient framework is highly required.

In this paper, we will introduce a DRL based comprehensive framework to jointly optimize computation offloading and resource allocation in the large-scale MEC system, with three key components installed, i.e., 2r-SAE, ASA and 2p-ER.

\section{System model and problem Formulation}
\subsection{System model}
\begin{figure}[htpb]
  \centering
	\includegraphics[width=8.8cm]{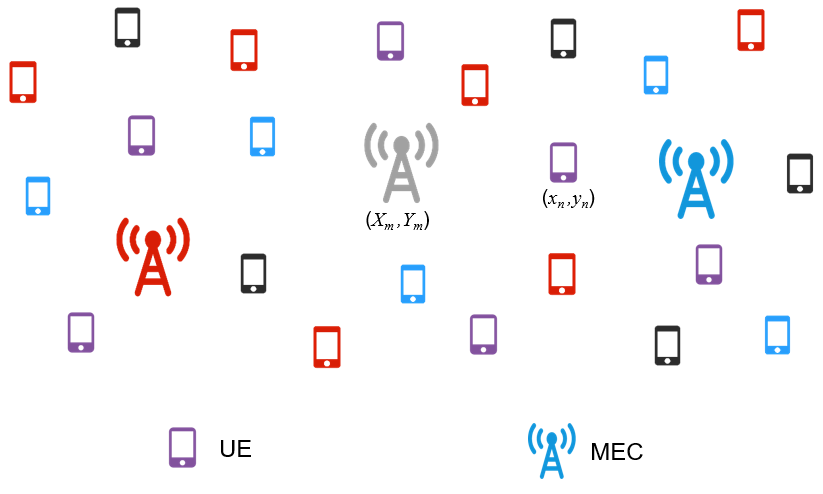}
	\caption{System model.}
  \label{fig:fig1}
\end{figure}
As shown in Fig. \ref{fig:fig1}, we consider there are $N$ UEs, denoted as a set of $\mathcal{N}=\{1,2,..., N\}$, each of which has a computational task to be executed. Also, we consider there are $M$ MECs, denoted as a set of $\mathcal{M}=\{1,2,\ldots,M\}$, which can enable UEs to offload their tasks. Define a new vector $\mathcal{M}^{\prime}=\{0,1,2,\ldots,M\}$ to denote the possible place which the tasks can be executed, therefore one has
\begin{align}\label{eq:Shi1}
 \mathrm{Cl}: a_{i j}=\{0,1\}, \forall i \in \mathcal{N}, \forall j \in \mathcal{M}^{\prime}
\end{align}
where $a_{i j}=1$, $j \neq 0$ denotes that the $i$-th UE decides to offload the task to the $j$-th MEC, while $a_{i j}=0$, $j\neq 0$ denotes that the $i$-th UE decides not to offload the task to the $j$-th MEC, and $a_{i j}=1$, $j=0$ denotes UE conducts the task itself. Also, one has 
\begin{align}\label{eq:Shi2}
 \mathrm{C} 2: \sum_{j \in \mathcal{M}^{\prime}} a_{i j} \leq 1, \forall i \in \mathcal{N}
\end{align}
which denotes that each task can only be or may not be able to execute in one place.

Similar to \cite{wang2016joint}, we assume that the $i$-th UE has the computational intensive task $U_{i}$ to be executed as
\begin{align}\label{eq:Shi3}
 U_{i}=\left(F_{i}, D_{i}\right), \forall i \in \mathcal{N}
\end{align}
where $F_{i}$ describes that the total number of the CPU cycles of $U_{i}$ to be computed, $D_{i}$ denotes the data size transmitting to the MEC if offloading action is decided. $D_{i}$ and $F_{i}$ can be obtained by using the approaches provided in \cite{yang2013framework}.

Then, one can have the execution time as
\begin{align}\label{eq:Shi4}
 T_{i j}^{C}=\frac{F_{i}}{f_{i j}}, \forall i \in \mathcal{N}, \forall j \in \mathcal{M}^{\prime}
\end{align}
where $f_{i j}$ is the computation capacity of the $j$-th MEC providing to the $i$-th UE and $j=0$ means the UE executes the task itself.

Then, the time to offload the data is given by \cite{zhu2019centralized}
\begin{align}\label{eq:Shi5}
 T_{i j}^{T r}=\frac{D_{i}}{r_{i j}}, \forall i \in \mathcal{N}, \forall j \in \mathcal{M}
\end{align}
where $r_{i j}$ is the offloading data rate from the $i$-th UE to the $j$-th MEC.

The computing capacity for the UE is constrained by
\begin{align}\label{eq:Shi6}
 \mathrm{C} 3: a_{i j} f_{i j} \leq F_{i, \ m a x }^{L}, \forall i \in \mathcal{N}, j=0
\end{align}
where $F_{i, \ m a x }^{L}$ is local computational capability of the $i$-th UE.

The power consumption of the UE is constrained by
\begin{align}\label{eq:Shi7}
 \mathrm{C} 4:  \sum_{j=1}^{M} a_{i j} p_{i j}^{T}+a_{i 0} p_{i}^{E} \leq P_{i, \ m a x }^{U E}
\end{align}
where $p_{i j}^{T}$ is the transmitting power from the $i$-th UE to the $j$-th MEC and $p_{i}^{E}$ is the execution power of the $i$-th UE if UE conducts the task itself. Thus, $p_{i}^{E}$ can be given by
\begin{align}\label{eq:Shi8}
 p_{i}^{E}=\kappa_{i}\left(f_{i j}\right)^{v_{i}}, \forall i \in \mathcal{N}, j=0
\end{align}
where $\kappa_{i} \geq 0$ is the effective switched capacitance and $v_{i} \geq 1$ is the positive constant. To match the realistic measurements, we set $\kappa_{i}=10^{-27}$ and $v_{i}=3$.

The computing capacity for the MEC is constrained by
\begin{align}\label{eq:Shi9}
\mathrm{C} 5: \sum_{i=1}^{N} a_{i j} f_{i j} \leq F_{j, \ m a x }^{M E C}, \forall j \in \mathcal{M}
\end{align}
where $F_{j, \ m a x }^{M E C}$ is the computational capability of the $j$-th MEC.

Assume that the coordinate of the $i$-th UE is $\left(x_{i}, y_{i}\right)$ and the coordinate of the $j$-th MEC is $\left(X_{j}, Y_{j}\right)$. The horizontal distance between the $i$-th UE and the $j$-th MEC is
\begin{align}\label{eq:Shi10}
 R_{i j}=\sqrt{\left(X_{j}-x_{i}\right)^{2}+\left(Y_{j}-y_{i}\right)^{2}}, \quad \forall i \in \mathcal{N}, \forall j \in \mathcal{M}.
\end{align}

Then, we can define channel states as
\begin{align}\label{eq:Shi11}
h_{i j}=\frac{\beta_{0}}{R_{i j}^{2}} l_{ij}, \forall i \in \mathcal{N}, \forall j \in \mathcal{M}
\end{align}
where $\beta_{0}$ denotes the channel power gain at the reference distance and $l_{ij}$ describes the influence of small-scale fading.

Therefore, if UEs decide to offload to the MEC, the data rate can be given as
\begin{align}\label{eq:Shi12}
r_{i j}=B \log _{2}\left(1+\frac{p_{i j}^{T} h_{i j}}{\sigma^{2}}\right), \forall i \in \mathcal{N}, \forall j \in \mathcal{M}
\end{align}
where $B$ is the channel bandwidth and $\sigma^{2}$ is the
noise spectral density.

\subsection{Problem Formulation}
In order to minimize the weighted sum of task latency of all the tasks, we formulate the optimization problem as follows:
\begin{displaymath}
P0: \min _{\mathbf{a}, \mathbf{f}, \mathbf{p}} \sum_{i \in \mathcal{N}} w_{i}\left(\sum_{j \in \mathcal{M}} a_{i j}\left(\frac{D_{i}}{r_{i j}}+\frac{F_{i}}{f_{i j}}\right)+a_{i 0} \frac{F_{i}}{f_{i 0}}\right)
\end{displaymath}
\begin{equation}\label{eq:Shi13}
\text { s.t. } \mathrm{C}1-\mathrm{C} 5
\end{equation}
where 
$\mathbf{a}=\left\{a_{i j}|i\in \mathcal{N}, j \in \mathcal{M}^{\prime}\right\}$, $\mathbf{f}=\left\{f_{i j}|i\in \mathcal{N},  j \in \mathcal{M}^{\prime}\right\}$ and $\mathbf{p}=\left\{p_{i j}|i\in \mathcal{N},  j \in \mathcal{M}^{\prime}\right\}$ are the vectors for offloading decision, computing resource allocation and transmission power of UEs, respectively. Also, one can see that it is a mixed-integer non-linear programming (MINLP), as it includes both integer and continuous variables. One can also see that if UE conducts the tasks locally, the energy consumption can be simply expressed as $p_{i 0}=p_{i}^{E}$. Also assume that $\mathbf{h}=\left\{h_{ij} | i \in \mathcal{N},  j \in \mathcal{M}\right\}$ is the time-varying variable, whereas other parameters are fixed values.


We first decompose $P$0 into two sub-problems, i.e., offloading decision sub-problem ($P$1), and transmission power and computation resource allocation sub-problem ($P$2). For $P$1, we assume that it only includes the integer variable $\mathbf{a}$, while other variables are fixed. Thus,  
one can see that $P$1 is an integer optimization, which is normally difficult to be solved in real-time under fast changing environment. To solve this issue, we propose to apply a novel DRL to address this problem and obtain the decision $\mathbf{a}$. Once $\mathbf{a}$ is obtained, $P$0 can be simplified as $P2$ as follows, with the integer variable $\mathbf{a}$ fixed.
\begin{displaymath}
P2: \min _{\mathbf{f}, \mathbf{p}} \sum_{i \in \mathcal{N}} w_{i}\left(\sum_{j \in \mathcal{M}} a_{i j}\left(\frac{D_{i}}{r_{i j}}+\frac{F_{i}}{f_{i j}}\right)+a_{i 0} \frac{F_{i}}{f_{i 0}}\right)
\end{displaymath}
\begin{equation}\label{eq:Shi14}
\text { s.t. } \mathrm{C3}-\mathrm{C} 5.
\end{equation}

One can see that the variables $\mathbf{p}$ can be set to its maximal value by applying $C4$. Then, $P$2 can be transformed to the minimization of the summation of fractional functions, which can be seen as the nonconvex sum-of-ratios optimization \cite{miettinen2010energy}.

 By applying
\begin{align}\label{eq:Shi15}
a_{i j}\left(\frac{D_{i}}{r_{i j}}+\frac{F_{i}}{f_{i j}}\right)+a_{i 0} \frac{F_{i}}{f_{i 0}} \leq \epsilon_{i j}
\end{align}
and combining Eq. (\ref{eq:Shi12}) and Eq. (\ref{eq:Shi15}), one can have
\begin{align}\label{eq:Shi16}
\mathrm{C} 6: D_{i}-B \log _{2}\left(1+\frac{p_{i j}^{T} h_{i j}}{\sigma^{2}}\right)\left(\frac{\left(\epsilon_{i j}-a_{i 0} \frac{F_{i}}{f_{i 0}}\right)}{a_{i j}}-\frac{F_{i}}{f_{i j}}\right) \leq 0.
\end{align}
Then, Problem $P$2 can be written as
\begin{displaymath}
P 2.1: \min _{\mathbf{f}, \mathbf{\epsilon}} \sum_{i \in \mathcal{N}} w_{i}\left(\sum_{j \in \mathcal{M}} \epsilon_{i j}\right)
\end{displaymath}
\begin{equation}\label{eq:Shi17}
\text { s.t. } \mathrm{C}3-\mathrm{C}6.
\end{equation}

One can see that $P$2.1 is a convex problem which can be solved by the standard convex optimization tool, e.g., CVX tool box.

\section{The Online joint resource scheduling framework (OJRS)}
Deep reinforcement learning (DRL) is a goal-oriented algorithm which can learn an optimal policy by using DNN for offloading decision making \cite{huang2019deep}.
In this paper, similarly, DRL is applied to predict computation offloading, i.e., $P$1, while convex optimization technique is used to solve $P$2 and evaluate the reward of DRL, which guarantees that all the physical constraints are satisfied. However, in a large-scale MEC system, there are three challenges for DRL to be directly applied: (1) because of the large number of UEs, the state space of DRL is extremely large, which increases the difficulty of policy learning; (2) the action search is very difficult because of the complex MINLP and the DRL is hard to find the best action and the learning process is inefficient; (3) the experience replay is sensitive to the environment, especially in dynamic situations, where the DRL is unstable and difficult to converge. These problems prohibit the DRL to be applied in the proposed problem \cite{henderson2018deep}. To address above challenges, we introduce an online joint resource scheduling (OJRS) framework which will be outlined next.
\subsection{The framework outline}

\begin{figure*}[htpb]
	\centering
	\includegraphics[width=18cm]{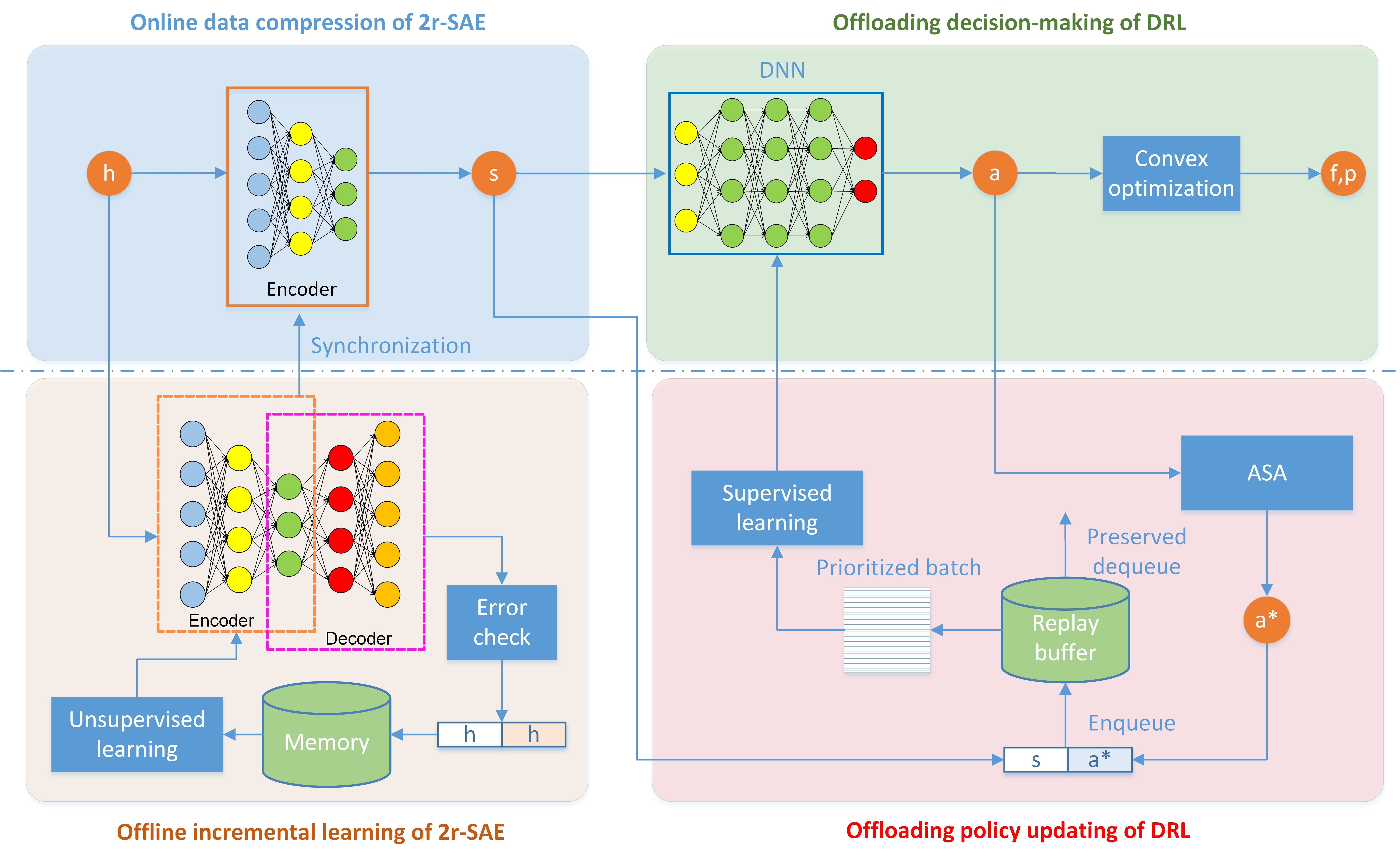}
	\caption{The OJRS framework.}
	\label{fig:fig2}
\end{figure*}

We show OJRS framework in Fig. \ref{fig:fig2}. There are three key improvements for solving the aforementioned problems: (1) the related and regularized stacked auto encoder (2r-SAE) is provided in Subsection-IV.B as a feature extractor, which can realize adaptive dimensionality reduction and data compression from the input, i.e., channel state information $\mathbf{h}$ by applying deep learning and hierarchical representation. The extracted feature is considered as the current state $\mathbf{s}$ of DRL, which is introduced in Subsection-IV.C; (2) an adaptive simulated annealing named ASA is presented in Subsection-IV.D as the heuristic search to help agent find better actions in DRL. Then the optimal offloading action $\mathbf{a}^{*}$ is achieved by maximizing the reward which is cached into the replay buffer of DRL; (3) a DNN is applied to devise the optimal offloading policy function $\pi$, which is achieved by a novel preserved and prioritized experience replay (2p-ER) in Subsection-IV.E. Finally, the convex optimization techniques is applied to solve the Problem $P$2.1 according to the given $\mathbf{a}$ and therefore the transmission power $\mathbf{p}$ and computation resource $\mathbf{f}$ can be calculated efficiently. The OJRS framework combines the hierarchical representation ability of deep autoencoder and the autonomous learning ability of DRL, which can realize an end-to-end online joint resource scheduling for large-scale MEC system in dynamic environment. The OJRS framework reduces the state space greatly by applying SAE. Meanwhile, the OJRS framework depends on no prior knowledge of environment, and can provide online decision making without solving the original MINLP problem. In the following, we provide the details of each component of the OJRS framework.

\subsection{2r-SAE}
An auto-encoder (AE) is a special and tricky feedback neural network with the same input and output by unsupervised learning. Consider the advantages of deep learning in feature extraction and representation learning, the SAE with multilayer encoder and decoder stacked by several AEs is shown in Fig. \ref{fig:fig2}, which assumes a symmetrical structure. Suppose the input vector $\mathrm{x} \in \mathbb{R}^{n}$, and the new representation $\mathrm{y} \in \mathbb{R}^{m}$, the encoder with $L$ layers describes a mapping:
\begin{align}\label{eq:Shi18}
\mathrm{x} \in \mathbb{R}^{n} \rightarrow \mathrm{r}_{L}=\mathrm{y} \in \mathbb{R}^{m}
\end{align}
where $\mathrm{r}_{L}$ is the output of the encoder through the iterative processing steps as follows:
\begin{align}\label{eq:Shi19}
r_{l}=f\left(r_{l-1} ; \theta_{l}\right)=\sigma\left(W_{l} r_{l-1}+b_{l}\right)
\end{align}
where $\mathrm{r}_{l} \in \mathbb{R}^{N_{l}}$ is the output of the $l$-th layer, $W_{l} \in \mathbb{R}^{N_{l} \times N_{l-1}}$ is the weight of the $l$-th layer, $b_{l} \in \mathbb{R}^{N_{l}}$ is the threshold of the $l$-th layer. The set of parameters for the $l$-th layer is $\theta_{l}=\left\{W_{l}, b_{l}\right\}$. $\sigma(\cdot)$ is the activation function which can be selected as sigmoid, tanh or ReLU\cite{8907406}. Then the decoder with $L$ layers describes a mapping:
\begin{align}\label{eq:Shi20}
\mathrm{y} \in \mathbb{R}^{m} \rightarrow \mathrm{r}_{2 L-1}=\hat{\mathrm{x}} \in \mathbb{R}^{n}
\end{align}
where $\hat{\mathrm{x}}$ is the reconstruction vector.


The SAE training aims to optimize the parameter set $\theta^{SAE}$, minimizing the reconstruction error between $x_{i}$ and $\widehat{x}_{i}$. The loss function of traditional SAE is always calculated as follows \cite{shao2017novel}:
\begin{align}\label{eq:Shi21}
L_{S A E}(\theta^{SAE})=\frac{1}{n} \sum_{i=1}^{n}\left(\frac{1}{2}\left\|x_{i}-\hat{x}_{i}\right\|^{2}\right)
\end{align}
where the mean square error (MSE) is usually used as the error term.

Gradient descent based methods are applied to tackle the loss minimization problem, i.e. iteratively updating the parameters $\theta^{SAE}$ according to the formula:
\begin{align}\label{eq:Shi22}
\theta^{SAE}(t+1)=\theta^{SAE}(t)-\beta \nabla L_{S A E}(\theta^{SAE}(t))
\end{align}
where $\beta$ is the learning rate, and $t$ is the iteration number.

SAE can be seen as a way to transform representation. When restricting the number of output nodes $m$ to be less than the number of original input nodes $n$ in the encoder, we can obtain a compressed representation of the input, which actually achieves desired dimensionality reduction. In large-scale MEC systems, the channel state matrix $\mathbf{h}$ is taken as the input vector for offloading decision making, and the input dimensionality of the $\mathbf{h}$ increases when the number of UEs and MECs are increased. Therefore, SAE can be used as a dimensionality reduction tool to hierarchically extract the key features of the original $\mathbf{h}$ and obtain a compact representation $\mathbf{s}$ as the input state of the DRL.

However, there are still two open problems in the design of SAE model for our problem: First, the error term of the loss function is MSE in SAE, which is an absolute error indicator for all UEs, but the relative CQI of each UE between different MECs provides key information for offloading decision. If we only consider absolute error in loss function, some UEs with small CQI values will have serious loss in the feature-extracting process. Second, the standard SAE only adopts MSE as the loss function, which is always prone to over-fitting and not suitable for online feature-extracting in our OJRS framework because of the poor generalization.

To address the above problems, we propose a novel related and regularized stacked auto encoder (2r-SAE) with an improved loss function, which can be implemented by
\begin{align}\label{eq:Shi23}
\begin{aligned}& L_{2r-SAE}(\theta^{SAE})= \frac{1}{NM} \sum_{i=1}^{N} \sum_{j=1}^{M}\left(\left\|h_{i j}-\hat{h}_{ij}\right\|^{2}\right) \\ &+\frac{\gamma_{1}}{2} \sum_{i=1}^{N} \sum_{j=1}^{M}\left\|\frac{h_{i j}}{\max \left\{h_{i k} | k \in \mathcal{M}\right\}}-\frac{\hat{h}_{i j}}{\max \left\{\hat{h}_{i k} | k \in \mathcal{M}\right\}}\right\|^{2} \\ &+\frac{\gamma_{2}}{2}\|\theta\|^{2} \end{aligned}
\end{align}
where $h_{i j}$ is the channel state information between the $i$-th UE and the $j$-th MEC, and the $\hat{h}_{ij}$ is the corresponding reconstruction output of SAE. In the loss function, the first term is the traditional absolute error term; the second term is the relative error term, which is used to maintain the relative size of $\mathbf{h}$ for each UE; and the third term is the regularized term, which is applied to improved generalization for online data compression.

In summary, as shown in Fig. \ref{fig:fig2}, the 2r-SAE is composed of two stages: (1) Offline incremental learning stage: In this stage, we introduce the SAE to preprocess the $\mathbf{h}$ matrix of all UEs and the unsupervised learning is used to extract the potential features of the $\mathbf{h}$ matrix and provide a compact state space for DRL, which will improve the robustness and efficiency of DRL in the large-scale MEC system. In addition, the incremental learning is used to train the SAE for tracking the variations of the real scenarios \cite{elwell2011incremental}. The procedure of incremental learning is described as follows. First, each $\mathbf{h}$ is input to the SAE, and a reconstruction error can be calculated. Then, we use an error check to decide if the current $\mathbf{h}$ can be put into the memory. In this paper, error check is a simple threshold evaluation, which means if the reconstruction error is larger than threshold, the current $\mathbf{h}$ will be put into the memory. Next, memory is a dynamic database with fixed-size, and first-in first-out (FIFO) scheduling policy is applied to the memory when the memory is full. Finally, the memory is used as the sample database to train the SAE. (2) Online data compression stage: The trained SAE can be implemented for online feature extraction and information compression. The extracted feature is considered as the current state $\mathbf{s}$ of DRL algorithm. The detailed description of 2r-SAE algorithm is provided in $\bf{Algorithm\enspace\ref{alg1}}$.

\begin{algorithm}
	\caption{2r-SAE algorithm}
	\label{alg1}
	\begin{algorithmic}[1]
		\REQUIRE   $\mathbf{h}$, $T_{S A E}$, $\gamma_{1}$,  $\gamma_{2}$.
		\ENSURE Compressed $\mathbf{s}$.
		\STATE{Rasterize CQI matrix $\mathbf{h}$ to a vector $\mathbf{x}$ as the input and label of SAE.}
		\STATE{Initialize the SAE network with random $\theta^{SAE}$.}
		\STATE{$Offline$ $incremental$ $learning$ $stage$:}
  	\STATE{Update memory by error check and select samples from memory.} 
		\WHILE{$t \leq T_{SAE}$}
		
		\STATE{Calculate the feedforward of SAE according to Eq. (\ref{eq:Shi19}) for all layers.} 
		\STATE{Calculate the loss function according to Eq. (\ref{eq:Shi23}).}			   
		\STATE{Update $\theta^{SAE}$ of SAE according to Eq. (\ref{eq:Shi22}).}
		\STATE{$t=t+1$.}
		\ENDWHILE\\
		\STATE{Synchronize the parameters from SAE  to the online encoder periodically.}
		\STATE{$Online$ $data$ $compression$ $stage$:}		
		\STATE{Calculate the output of encoder $\mathbf{s}$ based on the trained SAE according to the online input $\mathbf{h}$.}
	\end{algorithmic}
\end{algorithm}

\subsection{DRL with ASA and 2p-ER}
We use the other DNN to generate the optimal offloading action $\mathbf{a}$ of Problem $P$1 in real time, which can be regarded as an unknown function mapping $\pi $ from the compressed $\mathbf{s}$ to the optimal offloading action $\mathbf{a}$, namely:
\begin{align}\label{eq:Shi24}
\pi: \mathbf{s} \rightarrow \mathbf{a}.
\end{align}

However, it is challenging to collect sufficient number of labelled samples for DNN in practical MEC systems. Therefore, DRL is more suitable than supervised learning, as it can learn the offloading policy $\pi$ via the reward. By learning the offloading policy $\pi$ gradually from the interaction with environment, DNN can generate the best offloading decision behaviours by maximizing the rewards. Nevertheless, the traditional DRL cannot be directly applied for our problem due to the following two reasons: First, different from the traditional DQN, DNN in OJRS framework is used to directly generate actions instead of Q values and how to find the optimal action for improving the offloading policy $\pi$ remains unclear; Second, considering the dynamic environment, DRL is unstable and hard to converge, therefore a robust and efficient learning algorithm should be designed. 

Motivated by above issues, we propose a novel DRL, in which an ASA algorithm is applied to enhance the action search process and a 2p-ER strategy is used to improve the learning process of DNN. The schematic of the DRL is also illustrated in Fig. \ref{fig:fig2}. In the novel DRL algorithm, the agent interacts with the system environment in discrete decision epochs. At each epoch $t$, the agent carries out action $a_t$ according to the state $s_t$, then the environment produces a reword $r_t$ according to the action $a_t$. To improve the policy, a heuristic search is applied to search the optimal action $a_{t}^{*}$, and then the state-action pairs $\left\{h_t,a_{t}^{*} \right\}$ are put into the experience replay (ER) for agent learning. Concretely, in our problem, DNN can be seen as the agent, the $s_t$ is defined as the compressed $\mathbf{s}_{t}$ which is preprocessed by the 2r-SAE and acquired as the DNN’s inputs; the $a_t$ is defined as the offloading action $\mathbf{a}_{t}$ which is regarded as the DNN’s outputs; and the reward $r_t$ is deduced from the current $a_t$. For realizing the online decision-making process, we calculate $r_t$ directly by solving Problem $P$2.1 using convex optimization method which can be calculated efficiently and rapidly in the fast changing environment without considering the long-term reward. In addition, the reciprocal of the weighted task latency is defined as the reward of our DRL. The ASA is adopted as the heuristic search to find the optimal action for maximizing reward, and 2p-ER introduced in Subsection-IV.E is applied as the enhanced ER for DNN training in dynamic environment.

 In addition, different from the SAE, the offloading decision making is a classification task, thus the one-hot encoding is applied to transform the output of DNN to a specific category, and the regularized cross-entropy loss function of the DNN is selected as follows:
\begin{align}\label{eq:Shi25}
\begin{aligned}L\left(\theta_{t}\right)=&-\frac{1}{P} \sum_{i=1}^{P}\left(\left(a_{i}^{*}\right)^{T} \log \left(a_{i}\right)+\left(1-a_{i}^{*}\right)^{T} \log \left(1-a_{i}\right)\right)\\ 
&+\frac{\lambda}{2}\left\|\theta^\pi_{t}\right\|^{2} 
\end{aligned}
\end{align}
where $P$ is the sample set size; $a_i$ is the predicted offloading action from the DNN; $a_{i}^{*}$ is the labeled offloading action; and $\theta^\pi_{t}$ is the parameters of DNN at epoch $t$ which is updated by applying the Adam algorithm\cite{kingma2014adam} until the loss value is below a required threshold. Regularized term is also used in the loss function and the reasons are as follows: (1) regularized restraint will increase the generalization of DNN\cite{jiang2018electrical}; (2) the L2-norm of $\theta^\pi_{t}$ will record the status of DNN at each epoch which will be applied to preserve transitions in replay buffer.

\subsection{ASA}

 Action search plays a key role in our DRL, some local search methods are applied to find the best $a_{t}^{*}$ for improving the performance of DNN and achieving the optimal offloading policy $\pi $ \cite{huang2019deep}. However, these local search methods are easily stuck in local minima and the globally optimal offloading policy cannot be guaranteed. We introduce an adaptive simulated annealing (ASA) to carry out the global heuristic search for searching the best action $a_{t}^{*}$ and acquiring the optimal offloading policy $\pi$ in DRL. After heuristic search, the newly generated state-action pairs $\left\{s_t,a_{t}^{*} \right\}$ are appended to the replay buffer as training transitions of DNN.

Simulated annealing (SA) is a single-solution-based metaheuristic search inspired by the annealing in metallurgy. Due to its simplicity, less parameter, and fast convergence, SA has been widely adapted for global search and optimization during recent years\cite{zhang2018optimization}.

The traditional SA algorithm begins with an initial solution $x(0)$ and a starting temperature $T(0)$, then an iterative search process is carried out. For each generation $G$, a neighbor solution $x^{\prime}(G)$ close to the current solution $x(G-1)$ is generated by a randomly generation. The subsequent solution $x(G)$ is selected by the Boltzmann probability distribution\cite{zhang2018optimization}:
\begin{align}\label{eq:Shi28}
x(G)=\left\{\begin{array}{ll}{x^{\prime}(G)} & {\text { if } \exp \left(\frac{f(x(G-1))-f\left(x^{\prime}(G)\right)}{T(G)}\right)>r a n d} \\ {x(G-1)} & {\text { otherwise }}\end{array}\right.
\end{align}
where $f(\cdot)$ denotes the objective function of SA, $T(G)=\varphi T(G-1)$ which varies during the iterations because $\varphi \in (0,1)$ is the cooling factor. $rand$ denotes a uniform random number in the range [0, 1].

However, the traditional SA algorithm has three drawbacks that avoid its direct application in our DRL algorithm. Firstly, SA algorithm often employs continuous real-valued encodings, but the offloading decision $\mathbf{a}$ is a matrix with integer elements equal to 0 or 1; Second, traditional SA generates neighbour solutions randomly, and it does not take advantage of the CQI information; Third, the iteration number of SA is always fixed, which will lead to long computing time when the DRL finally converges. In this regard, we propose a new ASA algorithm to search the optimal action $\mathbf{a^{*}}$ efficiently.

First, we improve the coding of SA’s solution. In our ASA algorithm, the solution can be represented as:
\begin{align}\label{eq:Shi29}
\boldsymbol{a}=&[a_1,a_2,\cdots,a_i,\cdots,a_N]
\end{align}  
where $a_i=0$ means that the $i$-th UE decides to execute the task itself, and $a_i=k$ means that the $i$-th UE decides to offload the task to the $k$-th MEC, while $k \in \mathcal{M}$. This representation transforms the offloading decision matrix $\mathbf{a}$ to an integer coding for SA.

Second, channel quality $\mathbf{h}$ provides the prior information for guiding neighbour solution generation. We introduce an adaptive h-mutation to obtain the neighbour solution. The mutation probability of the $i$-th solution is given as:
\begin{align}\label{eq:Shi30}
P_{i}^{m u t}=\frac{h_{i, a_{i}}}{\sum_{j \in \mathcal{M}} h_{i j}}.
\end{align}
The adaptive h-mutation strategy is given as 
\begin{align}\label{eq:Shi31}
a_{i}^{\prime}=\left\{\begin{array}{cl}{{rand}m_{i}} & {\text { if }rand >P_{i}^{m u t}} \\ {a_{i}} & {\text { otherwise }}\end{array} \quad \forall i \in \mathcal{N}\right.
\end{align}
where ${randm}_{i} \in \mathcal{M}^{\prime}$  is a randomly generated integer to make sure that the $i$-th UE will offload the task to an MEC or execute the task itself. In the h-mutation strategy, the UE will have higher probability to offload the task to the MEC whose channel quality is better, so this strategy is better than random neighbour solution.

Third, $\Delta \delta_{t}$ of the DNN at each epoch $t$ is also introduced to adjust the iteration number $T_{SA}$ adaptively using the following equation:
\begin{align}\label{eq:Shi32}
T_{S A}(t+1)=\left\{\begin{array}{ll}{T_{S A}(t)+1} & {\text { if } \Delta \delta_{t} \geq \varepsilon} \\ {T_{S A}(t)-1} & {\text { if } \Delta \delta_{t}<\varepsilon \text { and } T_{S A}(t) \neq 1} \\ {1} & {\text { otherwise }}\end{array}\right.
\end{align}
where $\varepsilon$ is a threshold. In the adaptive iteration strategy, the iteration number of SA will decrease continuously in the training process of the DNN, while will increase when the environment varies, therefore this strategy is suitable for action search in dynamic environment and has high search efficiency.

Fourth, the convex optimization is applied to solve Problem $P$2.1 for each solution in ASA and 
Eq. (\ref{eq:Shi17}) is adopted as the objective function $f(\cdot)$. The detailed description of ASA algorithm is provided in $\bf{Algorithm\enspace\ref{alg3}}$.

\begin{algorithm}
	\caption{ASA algorithm}
	\label{alg3}
	\begin{algorithmic}[1]
		\REQUIRE   $a_t$, $\varphi$, $T_{SA}$, $\varepsilon$, $T(0)$ , $\Delta \delta_{t}$.
		\ENSURE  $a_{t}^{*}$.
		\STATE{Initialize $a_t$ as the $x(0)$.}
		\STATE{Update $T_{SA}(t)$ according to $\Delta \delta_{t}$ in Eq. (\ref{eq:Shi32}).}
		\WHILE{$G \leq T_{SA}(t)$}
		\STATE{Generate a neighbor solution $x^{\prime}(G)$  by Eqs. (\ref{eq:Shi30})-(\ref{eq:Shi31}).} 
		\STATE{ Calculate the fitness of the neighbour solution $x^{\prime}(G)$.}			   
		\STATE{Select subsequent solution $x(G)$ by Eq. (\ref{eq:Shi28}).}
		\STATE{Update $T(G)$.}
		\ENDWHILE\\
	\end{algorithmic}
\end{algorithm}

\subsection{2p-ER}


Experience replay (ER) is the other key technology in our DRL framework, because it has the following merits: (1) The random sampling can enhance stability of DRL by reducing the correlation between the samples in the buffer; (2) The reuse of history data can enhance the transition utilization and maintain the transition diversity, which will improve the performance of DNN\cite{van2016deep}. The procedure of ER is as follows: the buffer is empty at the beginning of the first epoch, and then the new state-action pairs $\left\{s_t,a_{t}^{*} \right\}$ at the epoch $t$ are collected and added to the buffer. Next, the random batch sampling in the buffer is applied to train DNN, and new transitions will be collected from the trained DNN continually. When the buffer is full, FIFO scheduling policy is employed, and the oldest transitions will be discarded. However, traditional ER may discard some good transitions when the buffer is full because of the FIFO strategy, and the selection probability of all transitions is uniform. These traits limit the learning efficiency of DNN, especially in the dynamic environment. To address these obstacles, we propose a preserve strategy and a priority strategy in replay buffer whose details are described as follows:

(1) Preserve strategy: in replay buffer, we will preserve the transitions which are similar to the current offloading policy $\pi_{t}$. During the training process, the offloading policy gradually shifts away from the previous status, and the samples whose offloading policy are different from the current offloading policy may not contribute to DNN’s outcomes. The difference between the offloading policy of the transition $i$ collected at epoch $t^{\prime}$ and current offloading policy $\pi_{t}$ can be measured as follows:
\begin{align}\label{eq:Shi26}
\rho_{i}=\frac{\left\|\theta^\pi_{t}\right\|^{2}}{\left\|\theta^\pi_{t^{\prime}}\right\|^{2}}
\end{align}
where $\left\|\theta^\pi_{t}\right\|^{2}$ is the L2-norm of ${\theta^\pi_{t}}$ at the current epoch $t$, and $\left\|\theta^\pi_{t^{\prime}}\right\|^{2}$ is the L2-norm of $\theta^\pi_{t^{\prime}}$ at the epoch $t^{\prime}$ which is the transition collected epoch. Thus we compute a dissimilarity factor of each transition and define the reusable transition if $\frac{1}{\rho_{\ m a x }}<\rho_{i}<\rho_{\ m a x }$ with $\rho_{\ m a x }>1$. The reusable transitions will be preserved and reused during the FIFO process.

(2) Priority strategy: in replay buffer, the transition which incurs obvious loss function decrease will be set with the higher selection probability, while the transition which cannot improve the performance of DNN obviously will be set with the lower selection probability. This strategy will increase the learning frequency of the valuable transitions and eliminate inefficiencies in the DRL process. The probability of sampling transitions $i$ is defined as:
\begin{align}\label{eq:Shi27}
P_{i}=\frac{p_{i}^{\tau}}{\sum_{k \in \mathcal{K}} p_{k}^{\tau}}
\end{align}
where $p_{i}=\left|\Delta \delta_{t}\right|+\epsilon$, $\epsilon$ is a small positive constant which guarantees that all the transitions can be sampled, even if the variation of loss function $\Delta \delta_{t}=0$ at epoch $t$\cite{schaul2015prioritized}. In (\ref{eq:Shi27}), $\mathcal{K}$ is the set of all transitions in the replay buffer and $\tau$ is a probability factor to control how much priority is used.

To realize the preserve and priority strategy in replay buffer, we sort two extra variable $\left\{\Delta \delta_{t},\left\|\theta^\pi_{t}\right\|^{2}\right\}$ at epoch $t$ when we update the DNN. It is worth noting that these two strategies are readily to process because $\delta_{t}$ and $\left\|\theta^\pi_{t}\right\|^{2}$ have been calculated at the loss function already.

In summary, as shown in Fig. \ref{fig:fig2}, the DRL with ASA and 2p-ER is composed of two alternating stages: (1) Offloading decision making stage: At epoch $t$, the DNN whose parameters are represented as the offloading policy $\pi_{t}$ can be deployed for generating online offloading action $a_t$ according to $s_t$, then the convex optimization algorithm is used to solve $P$2.1 and calculate $p_t$ and $f_t$ according to $a_t$, which guarantees that all constraints are satisfied. Then the solutions $\left\{a_t, p_t, f_t\right\}$ for $h_t$ can be output in real time; (2) Offloading policy updating stage: The computation offloading $a_t$ is set as the initial solution of the ASA search. Then the ASA search is introduced to improve the action $a_t$ and the best $\left\{h_t,a_{t}^{*} \right\}$ is selected as the new transition and appended to the replay buffer. After that, a batch of transitions are drawn from the buffer according to our preserve and priority strategy, and the DNN is trained and the offloading policy is updated from $\pi_{t}$ to $\pi_{t+1}$. Meanwhile the variable $\left\{\Delta \delta_{t},\left\|\theta^\pi_{t}\right\|^{2}\right\}$ is recorded to update the $\rho_i$ and $P_i$ of selected transitions. The new offloading policy $\pi_{t+1}$ is applied in the epoch $t+1$ to generate the offloading decision $a_{t+1}$ according to the new $s_{t+1}$. These two stages are alternatively performed and the offloading policy is gradually improved in the iteration process. The detailed description of DRL with ASA and 2p-ER is provided in $\bf{Algorithm\enspace\ref{alg2}}$.

\begin{algorithm}
	\caption{DRL with ASA and 2p-ER}
	\label{alg2}
	\begin{algorithmic}[1]
		\REQUIRE  $h_t$, $\tau$, $\rho_{max}$, $\tau$, $T_{DRL}$, training interval $\phi$.
		\ENSURE $a_t$.
		\STATE{Initialize the DNN with random $\theta^\pi_{0}$.}
		\STATE{Initialize an empty replay buffer.}
		\WHILE{$t<T_{DRL}$}
		\STATE{Generate the offloading action $a_t$ according to the offloading policy $\pi_{t}$.}
		\STATE{Find the best $a_{t}^{*}$ by \bf{Algorithm\enspace\ref{alg3}}.}
		\STATE{Append the state-action pair $\left\{h_t,a_{t}^{*} \right\}$ to the replay buffer.}			   
		\STATE{Protect the reusable transitions by preserve strategy if the buffer is full.}
		\IF{$t \bmod \phi=0$}
		\STATE{Sample a batch of transitions by priority strategy.}
		\STATE{Train the DNN and update the offloading policy using the loss in Eq. (\ref{eq:Shi25}).}
		\STATE{Record $\left\{\Delta \delta_{t},\left\|\theta^\pi_{t}\right\|^{2}\right\}$ and update $\rho_i$ and $P_i$ of selected transitions.}
		\ENDIF
		\ENDWHILE
	\end{algorithmic}
\end{algorithm}

\section{Numerical results and discussion}
\subsection{Simulation parameters setting}
Our simulation parameters are given in TABLE \ref{tab:table1}, unless otherwise specified. The parameters of the 2r-SAE are chosen as follows: We adopt a 3-layer fully-connected feedforward neural network to serve as the encoder of SAE, which includes 60, 45 and 30 neurons in the first, second and third layers respectively as well as $T_{SAE}$=500, $\gamma_{1}$=0.5 and $\gamma_{2}$=0.08. The parameters of the DRL are chosen as follows: We use a 4-layer fully-connected feedforward neural network to serve as the DNN, which includes 30, 120, 80 and 30 neurons in each layer respectively, as well as $\lambda$=0.02, $T_{DRL}$=10000 and $\phi$=10. The parameters of the ASA are chosen as follows: $T_{SA}$=20 and $\varepsilon$=0.02. The parameters of the 2p-ER are chosen as follows: $\rho_{max}$=1.2 and $\epsilon=0.001$. For Section V.B and V.C, we assume there are two MEC servers with coordinates (10m,10m) and (40m,40m)
located in the areas with squared size 50m*50m. For Section V.D, we vary the number of the MEC servers from one to five as follows: The locations of 1 MEC, 2 MECs, 3 MECs, 4 MECs and 5 MEC are respectively assumed as [(25m, 25m)]; [(10m, 10m), (40m, 40m)]; [(10m, 10m), (25m, 25m), (40m, 40m)]; [(10m, 10m), (10m, 40m), (40m, 10m), (40m, 40m)] and [(10m, 10m), (10m, 40m), (25m, 25m), (40m, 10m), (40m, 40m)]. Also, we assume there are 30 UEs, randomly distributed in the above area. 

\begin{table}[]
	\centering\makegapedcells
	\caption{Simulation parameters}
	\label{tab:table1}
	\begin{tabular}{|l|l|}
		\hline
		$\bf{Parameters}$  & $\bf{Assumptions}$ \\ \hline
		Data size of task  ${D_i}$  & 100kB \\ \hline
		Required CPU cycles of task  ${F_i}$  & $10^9$ cycles/s \\ \hline
		Bandwidth $B$  & 1MHz \\ \hline
		Local Computational Capability $F_{max}^{L}$  & $10^9$ cycles/s \\ \hline
		Remote Computational Capability $F_{max}^{MEC}$  & $50 \cdot 10^9$ cycles/s \\ \hline
	\end{tabular}
\end{table}

\subsection{2r-SAE performance evaluation}

\begin{figure}[htpb]
	\centering
	\includegraphics[width=8.8cm]{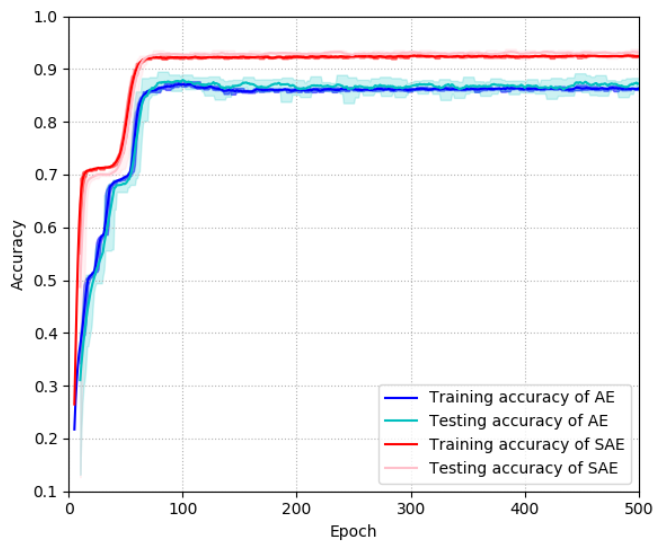}
	\caption{Comparison of prediction accuracy for AE and SAE.}
	\label{fig:fig5}
\end{figure}

\begin{figure}[htpb]
	\centering
	\includegraphics[width=8.8cm]{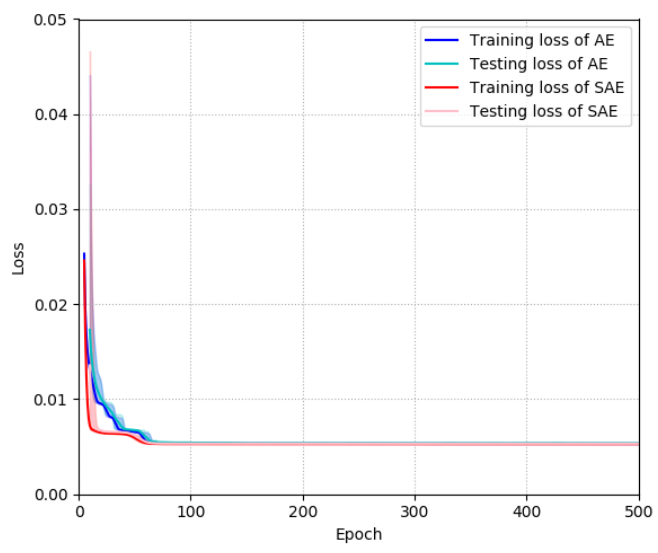}
	\caption{Comparison of loss for AE and SAE.}
	\label{fig:fig15}
\end{figure}

2r-SAE can provide a compact data representation to the DRL model. Fig. \ref{fig:fig5} characterizes the reconstruction accuracy of AE and SAE for the data compression and representation of channel state $\mathbf{h}$ in the MEC system with 2 MEC servers. The encoder of AE is a simple 2-layer fully-connected feedforward neural network, which includes 60 and 30 neurons in the first and second layers, respectively. It can be observed that the reconstruction accuracy of SAE is 92.73\% while the reconstruction accuracy of AE is 87.55\%. The SAE with 3 layers has more precise representation than traditional AE with 2 layers. This is due to the fact that the depth of the DNN directly affects the potential feature representation and extraction of $\mathbf{h}$ which in turn directly affects the reconstruction accuracy. Additionally, in Fig. \ref{fig:fig15}, the training losses of AE and SAE all converge to 0.0055 around after about 80 episodes, while the same phenomenon can be observed in testing loss curves, which means the unsupervised learning of AE and SAE can be used in channel states data preprocessing and compression successfully and the overfitting can be avoid.

\begin{figure}[htpb]
	\centering
	\includegraphics[width=8.8cm]{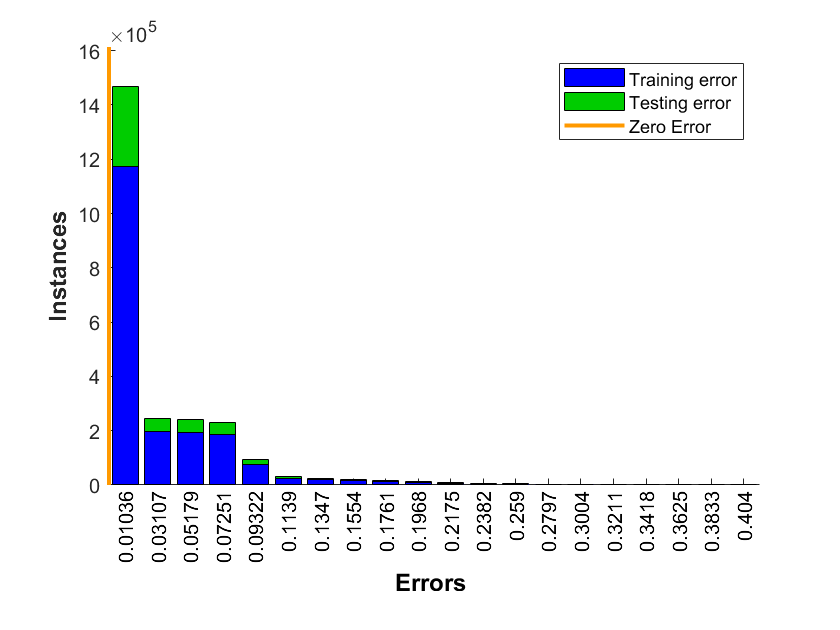}
	\caption{Absolute error distribution for AE.}
	\label{fig:fig6}
\end{figure}

\begin{figure}[htpb]
	\centering
	\includegraphics[width=8.8cm]{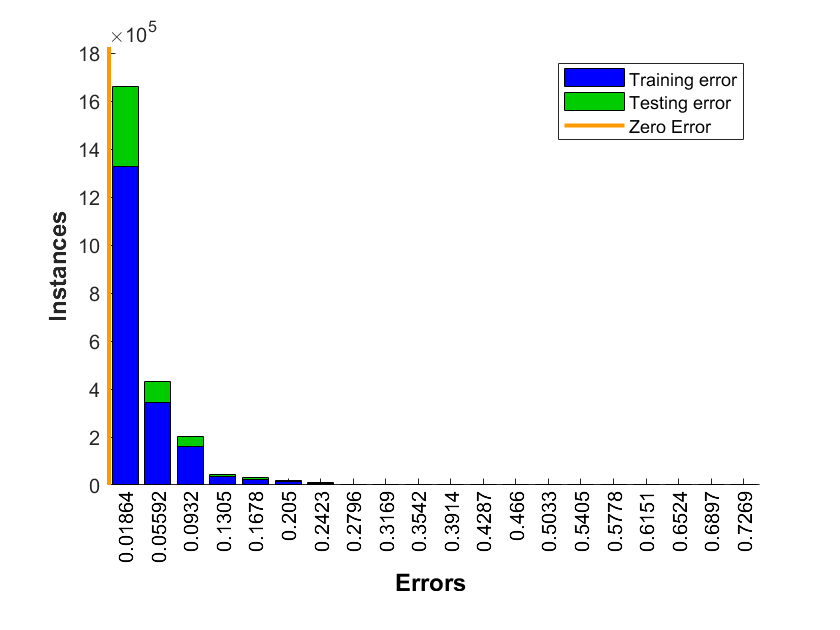}
	\caption{Absolute error distribution for SAE.}
	\label{fig:fig16}
\end{figure}

Fig. \ref{fig:fig6} and Fig. \ref{fig:fig16} characterize the absolute error distribution of all channel state data for 2r-SAE and standard SAE. The error values are chosen according to the statistics of samples.
One can see that the training error and testing error of 2r-SAE are more focused at the minimal error bar. There are two reasons to explain this phenomenon: Firstly, the relative error loss term of each UE is added to the loss function, so that the SAE considers not only MSE, but also the relative error of each data in the training process, which leads to the lower training error. Secondly, the regularized term ensures the generalization of SAE, which leads to the lower testing error.

\subsection{DRL performance evaluation}
\begin{figure}[htpb]
	\centering
	\includegraphics[width=8.8cm]{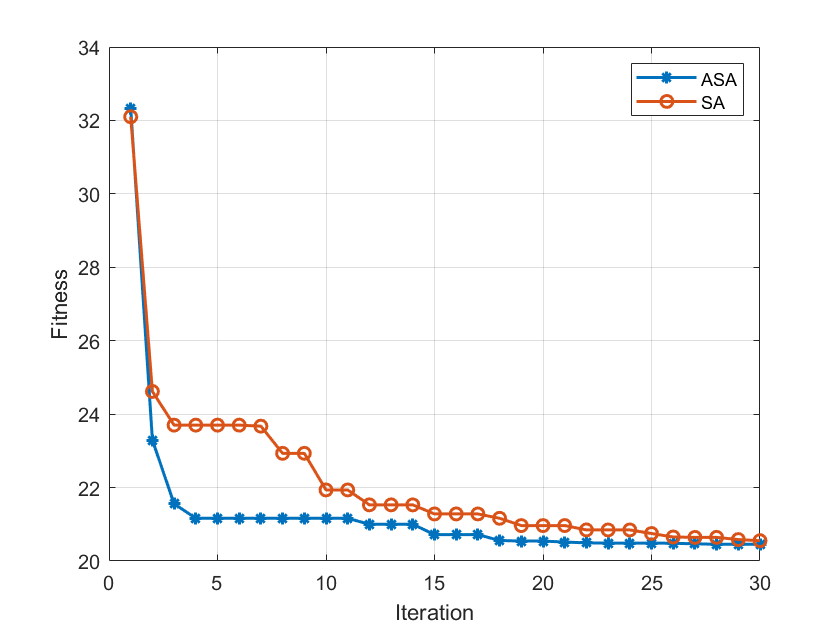}
	\caption{The action search process using ASA and the traditional SA.}
	\label{fig:fig10}
\end{figure}

\begin{figure}[htpb]
	\centering
	\includegraphics[width=8.8cm]{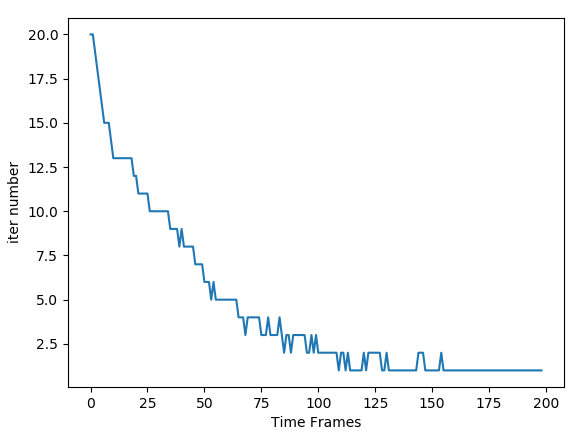}
	\caption{The adaptive iteration number of ASA during the DRL stage.}
	\label{fig:fig11}
\end{figure}

ASA is a key element to affect the performance of our DRL. Fig. \ref{fig:fig10} characterizes the best fitness values during the action search process using ASA and the traditional SA. It is observed that the ASA achieves the optimal action with less iterations and higher efficiency than SA. This is because h-mutation is applied to guide the action search and prompt the ASA to find the optimal neighbor solution efficiently. Fig. \ref{fig:fig11} characterizes the adaptive iteration number of ASA during the DRL stage. We can see that the iteration number of ASA decreases to 1 with the decline of $\Delta \delta_{t}$. At some special DRL epochs, the iteration number of ASA increases because of the augment of $\Delta \delta_{t}$. The adaptive iteration number will reduce the times of solving the convex optimization problem and further improve the computational efficiency of DRL.

\begin{figure}[htpb]
	\centering
	\includegraphics[width=8.8cm]{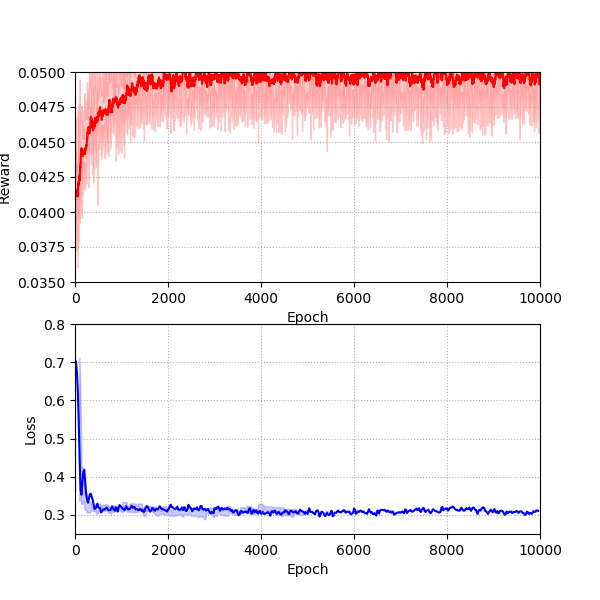}
	\caption{The reward and the loss value for DRL with 2p-ER.}
	\label{fig:fig8}
\end{figure}

\begin{figure}[htpb]
	\centering
	\includegraphics[width=8.8cm]{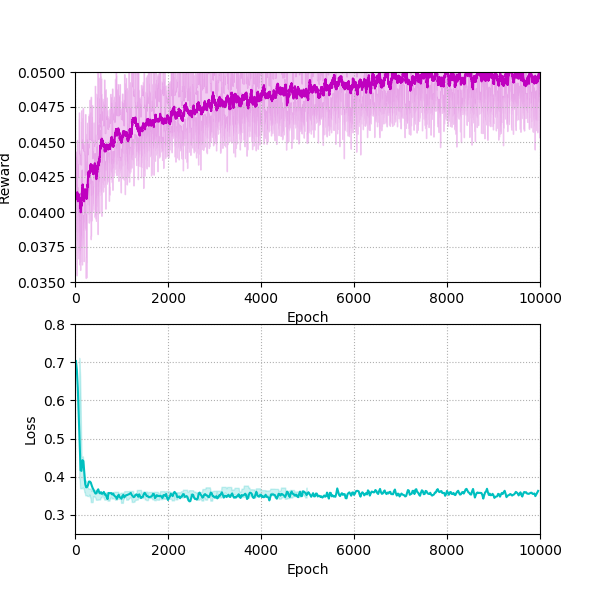}
	\caption{The reward and the loss value for DRL with traditional replay buffer.}
	\label{fig:fig9}
\end{figure}

2p-ER is another element to affect the performance of our DRL. Fig. \ref{fig:fig8} characterizes the reward and the loss value for our DRL with 2p-ER, while Fig. \ref{fig:fig9} characterizes the reward and the loss value for DRL with  $\epsilon$-greedy and traditional replay buffer. We can see that, for both offloading policies learned from DRL with 2p-ER and traditional DRL, the reward of each epoch increases as the interaction between the DNN and the MEC system environment continues, which indicates that DRL can acquire efficient offloading policies successfully without any prior environment knowledge. Besides, the reward of our DRL becomes stable after about 2500 epochs, while the reward of traditional DRL becomes stable after about 7000 epochs. On the other hand, loss performance of the DNN (offloading policy) learned from our DRL is always lower than traditional DRL. This is because the preserve strategy preserves the reusable transitions and enhances the correlation between the transitions and the current offloading policy. In addition, the priority strategy makes the transitions which can lead to the decline of loss function have higher selection probability. All of the above strategies improve the performance of 2p-ER.

\subsection{OJRS framework performance evaluation}

\begin{table}[]
	\centering\makegapedcells
	\caption{The performance comparison of scheduling strategies.}	
	\label{tab:table2}
	\begin{tabular}{|p{65pt}<{\centering}|p{45pt}<{\centering}|p{45pt}<{\centering}|p{45pt}<{\centering}|}
		\hline 
		Metric & Computational time (Sec) & Task latency (Sec) & Reward \\ 
		\hline 
	OJRS framework & 0.0174 & 20.6874 & 0.0483  \\ 
		\hline 
		Greedy & 0.0139 & 25.2942 & 0.0395 \\ 
		\hline 
		Random & 0.0057 & 36.4325 & 0.0275 \\ 
		\hline 
	ASA & 0.2354 & 20.2385 & 0.0494 \\ 
		\hline 
	\end{tabular}
\end{table}

Finally, we evaluate the whole OJRS framework. TABLE \ref{tab:table2} characterizes the performance of the proposed OJRS framework for online joint resource scheduling. The Greedy, Random and ASA are used as the benchmarks. Random offloading (Random) denotes that the offloading admission is decided randomly for each UE. If the computational resource of the allocated MEC is insufficient, UE executes the task locally. Greedy offloading (Greedy) denotes that all UEs offload the task to the nearest MEC. If the computational resource is insufficient, the UEs who need more computing resources execute the task locally. ASA denotes that the task offloading decision is optimized by the ASA method directly, without applying DRL. It can be observed that the ASA achieves the highest reward. The proposed method attains almost the same reward compared with ASA, which is higher than Greedy and Random. This is because the proposed method uses ASA to search the action space and constructs an optimal non-linear offloading policy from compressed $\mathbf{s}$ to offloading decision $\mathbf{a}$. Meanwhile, if the SAE and DRL are applied, the complexity of the proposed method in online decision making is far lower than that of the ASA.

TABLE \ref{tab:table3} characterizes the performance of the proposed OJRS framework in dynamic environment. We compare the accuracy and compression ratio of SAE with different number of MECs, and we also compare the best reward and average reward acquired from the DRL with varying weights. Especially, we consider a constant number of output neurons in SAE which is set to 30 when the number of MECs is changed, and we also consider a random weight variation at the 5000th epoch for simulating the dynamic environment. In order to evaluate the performance of DRL in different scenarios, we define the normalized reward rate (NRR), which is equal to that the inferred reward dividing the optimal reward. In NRR, the inferred reward in the numerator is obtained from the offloading decision of the DNN, and the optimal reward in the denominator is obtained from the particle swarm optimization (PSO) which is suitable for solving large-scale MINLP problems and can normally achieve nearly optimal global solutions but with long computation time \cite{guo2018efficient}.

The data such as the  accuracy of SAE (Acc), the compression ratio of SAE (CR), the best NRR (F-Best) and the average NRR (F-Avg) of DRL before the 5000th epoch, and the best NRR (S-Best) and the average NRR (S-Avg) of DRL after the 5000th epoch are saved in Table \ref{tab:table3} for detailed statistical analysis.

It can be observed that the reconstruction accuracy of SAE decreases when the number of MECs increases, while the compression ratio of SAE increases when the number of MEC server increases. Therefore if we are willing to accept some loss of reconstruction accuracy, we can obtain a larger compression ratio, especially for a large-scale MEC system. 

It also can be inferred from the results that the NRR of DRL also decreases when the number of MECs increases because of the information loss of SAE. However, this loss is compensated by the large compression ratio for state space, which will lead to fast search ability and stable convergence speed of DRL. Moreover, the DRL before the 5000th epoch achieves the same best NRR compared with the DRL after the 5000th epoch achieves, which means the proposed DRL can adjust the offloading policy automatically and it is suitable for making offloading decisions in dynamic environment. The average NRR of the DRL before the 5000th epoch is higher than the DRL after the 5000th epoch. A possible explanation of this phenomenon is that when the weights are changed, the DRL should just adjust the offloading policy to adapt the new environment, which is easier than the learning process of the original DRL without any prior information.

\begin{table}[]
	\centering\makegapedcells
	\caption{The performance of OJRS framework in dynamic environment.}	
	\label{tab:table3}
\begin{tabular}{|p{17pt}<{\centering}|p{25pt}<{\centering}|p{25pt}<{\centering}|p{25pt}<{\centering}|p{25pt}<{\centering}|p{25pt}<{\centering}|p{25pt}<{\centering}|}
	\hline
	\multirow{3}{*}{} & \multicolumn{6}{l|}{Performance}                         \\ \cline{2-7} 
MEC No.	& \multicolumn{2}{l|}{SAE} &  \multicolumn{4}{l|}{DRL} \\ \cline{2-7} 
	&      Acc     &    CR       &  F-Best   &   F-Avg  &  S-Best   &   S-Avg   \\ \hline
1	&    1       &    0       &  0.9987   &  0.9462   & 0.9988    &  0.9764   \\ \hline
2	&    0.9273       &   0.5        &   0.9895  &  0.9421   &   0.9886  &   0.9693  \\ \hline
3	&     0.8994      &    0.67       &   0.9821  &   0.9362  &   0.9823  &  0.9612   \\ \hline
4	&        0.8823   &    0.75       &   0.9732  &  0.9252   &  0.9733   &  0.9575   \\ \hline
5	&      0.8782     &   0.80        &  0.9672   &    0.9197 &  0.9671   &   0.9488  \\ \hline
\end{tabular}
\end{table}



\section{Conclusion}
In this paper, we have proposed a DRL based online joint resource scheduling framework. This framework adopts a SAE and a DRL to optimize computation offloading, transmission power, and computation resource in a large-scale MEC system. More particularly, a novel 2r-SAE with unsupervised learning is presented to carry out data compression and representation for high dimensional channel state data, which can reduce the state space of DRL. Secondly, a novel DRL is proposed to make offloading decision, in which an ASA is used to search the optimal action and a 2p-ER is used to assist the DRL to train the DNN and find the optimal offloading policy. Specifically, the ASA uses adaptive h-mutation and iteration to enhance the action search and further improve the computing efficiency during the DRL process. In addition, the 2p-ER applies preserve and priority strategies to optimize the ER and improve the training process of DNN. It is demonstrated that the proposed framework is capable of optimizing the computation offloading and resource allocation jointly at a high accuracy, making real-time resource scheduling feasible for large-scale MEC systems.

The future work will focus on the following aspects: 1) How to apply the proposed framework into the real-world systems, i.e., we will consider the hardware constraints and the real-world datasets; 2) How can we theoretically analyse the convergence of DRL-based algorithm in multi-user multi-MEC environment; 3) We can consider to extend this work by applying distributed DRL framework, such as MADDPG \cite{10.5555/3295222.3295385}, in order to improve the performance and further enhance security and privacy for each user.


%

%




\bibliographystyle{ieeetran}
\bibliography{bare_jrnl_bobo}

\end{document}